%% file: paper.tex
\documentclass[11pt, fullpage]{article}

\usepackage{url} 
\usepackage{float}
\usepackage{graphicx}
\usepackage{amsfonts, amssymb, amscd}
\usepackage{amsmath, delarray, theorem}
\usepackage{natbib}
\bibpunct{(}{)}{;}{a}{}{,}
\renewcommand{\cite}{\citep}

\usepackage{myfile} 

\newcommand{\mycap}[1]{\caption{{#1}}}
\newcommand{\bitem}{\begin{itemize}}
\newcommand{\eitem}{\end{itemize}}
\newcommand{\benum}{\begin{enumerate}}
\newcommand{\eenum}{\end{enumerate}}
\newcommand{\beqnn}{\begin{eqnarray*}}
\newcommand{\eeqnn}{\end{eqnarray*}}
\newcommand{\beqn}{\begin{eqnarray}}
\newcommand{\eeqn}{\end{eqnarray}}

\newcommand{\myvec}[1]{\mathbf{#1}}

\newcommand{\bfourmatrix}{\begin{array}({cccc})}
\newcommand{\efourmatrix}{\end{array}}
\newcommand{\bvector}{\begin{array}({c})}
\newcommand{\evector}{\end{array}}

\newcommand{\btab}{\renewcommand{\baselinestretch}{1}\begin{table}[hptb]}
\newcommand{\etab}{\end{table}\renewcommand{\baselinestretch}{1.75}}
\newcommand{\bfig}{\renewcommand{\baselinestretch}{1}\begin{figure}[hptb]}
\newcommand{\efig}{\end{figure}\renewcommand{\baselinestretch}{1.75}}

{\theoremstyle{plain}
  }
{\theoremstyle{plain}
  }
{\theoremstyle{plain}
  }
{\theorembodyfont{\rmfamily}
  }
{\theorembodyfont{\rmfamily}
  }

\renewcommand{\baselinestretch}{1.75}

\begin{document}

\def\mytitle{%
Classifying Network Data with Deep Kernel Machines
}

\def\myabstract{%
Inspired by a growing interest in analyzing network data, we study the 
problem of node classification on graphs, focusing on approaches based on 
kernel machines. Conventionally, kernel machines are linear classifiers in 
the implicit feature space. We argue that linear classification in the 
feature space of kernels commonly used for graphs is often not enough to 
produce good results. When this is the case, one naturally considers 
nonlinear classifiers in the feature space. We show that repeating this 
process produces something we call ``deep kernel machines.'' We provide 
some examples where deep kernel machines can make a big difference in 
classification performance, and point out some connections to various 
recent literature on deep architectures in artificial intelligence and 
machine learning. }

\def\mykeywords{%
deep architecture;
diffusion kernel;
kernel density estimation;
nearest centroid;
social network;
support vector machine.
}

\title{\mytitle}
\author{Xiao Tang and Mu Zhu\\
Department of Statistics and Actuarial Science \\
University of Waterloo \\
Waterloo, Ontario, Canada N2L 3G1}
\date{\today}
\maketitle

\begin{center} 
{\bf\large Abstract}
\end{center}  

\vspace{0.5cm}

{\small \myabstract}

\vspace{0.5cm}

\noindent
{\small {\bf Key Words}: \mykeywords}

\pagebreak
\parindent=0.75cm
\parskip=2mm
\input{main}

\section*{Acknowledgment}

This research is partially supported by the Natural Science and 
Engineering Research Council (NSERC) of Canada.

\bibliographystyle{/u1/m3zhu/natbib}
\bibliography{/u1/m3zhu/mzstat}

\end{document}

%% file: main.tex
\section{Introduction}

\def\cG{\mathcal{G}}
\def\cD{V_{obs}(\cG)}
\def\cDn{V_{miss}(\cG)}

Given a graph, $\cG$, let $V(\cG) = \{\myvec{v}_1, \myvec{v}_2, ..., 
\myvec{v}_n\}$ denote its set of vertices (nodes). Associated with each 
node $\myvec{v}_i$ is a class label, say $y_i \in \{1, 2, ..., 
C\}$. A small number of these labels are known; others are unknown. We use 
the notation
\[
 \cD =  \{\myvec{v}_i \in V(\cG): y_i \mbox{ observed}\} 
 \quad\mbox{and}\quad
 \cDn = \{\myvec{v}_i \in V(\cG): y_i \mbox{ missing}\}
\]
to denote subsets of nodes whose labels are known and unknown, 
respectively. 
The problem that we study in this article is that of predicting the 
unknown labels based on how the nodes are connected to each other. 

For example, $\cG$ may be a protein interaction network. 
Some proteins are known to be associated with certain biological functions 
(e.g., cell communication), and we may be interested in predicting which 
of the remaining proteins are also associated. Another example of $\cG$ is 
a social network, where the nodes are individuals. Some individuals are 
known to be involved with certain activities (e.g., belonging to a certain 
club, or interested in certain products), and we may wish to assess the 
likelihood that the remaining individuals are also involved.
For simplicity, we will assume $C=2$, i.e., there are only two classes, 
but our ideas also apply when $C > 2$. 

We develop something we call ``deep kernel machines'' (DKMs) on graphs. 
Though we focus on the problem of node classification on graphs, it will 
become clear by the end of this article (Section~\ref{sec:disc}) that the 
idea of DKMs does not really depend on the graph structure of the data. 
The main reasons why we choose to start with the node classification 
problem on graphs are: (a) it is a relatively new problem and not as 
widely studied as some of the more conventional classification problems 
\citep{eric-net}; and (b) it is a problem for which kernel machines are 
particularly well suited \citep{kernel-bk}. Experiments with data 
structures other than graphs will be left as future work.

\section{Kernel machines and graphs}
\label{sec:km}

\newcommand{\iprod}[2]{\langle #1, #2 \rangle}
\newcommand{\iprodv}[2]{\iprod{\myvec{v}_{#1}}{\myvec{v}_{#2}}}
\newcommand{\iprodphiv}[2]{\iprod{\phi(\myvec{v}_{#1})}{\phi(\myvec{v}_{#2})}}

Lately, largely due to support vector machines \citep[see, 
e.g.,][]{vapnik, svmbk}, kernel machines 
have become very popular in machine learning \citep{kernel-bk, 
bishop-mlbk}. Typically, a kernel machine has the form, 
\beqn 
\label{eq:km}
 f(\myvec{v}) = \alpha_0 + 
 \sum_{\myvec{v}_i \in \cD} \alpha_i \myvec{K}(\myvec{v}, \myvec{v}_i),
\eeqn
where $\myvec{K}(\cdot,\cdot)$ is a kernel function, and
the coefficient $\alpha_i$ often depends on the class label $y_i$. 

A support vector machine (SVM) is sometimes called a sparse kernel machine 
because the optimization problem it solves will cause many of the 
$\alpha_i$'s to be $0$. As a result, the decision function (\ref{eq:km}) 
depends only on those observations $(\myvec{v}_i, y_i)$ with $\alpha_i > 
0$, and hence the name ``sparse.''
In order to crystalize the gist of our ideas, we shall mostly work with a 
much simpler kernel machine (Section~\ref{sec:simple-km}), 
but other kernel machines such as SVMs also can be used. 

As noted by \citet{kernel-bk}, kernel machines are particularly well 
suited to analyze non-vectorial data such as graphs. Before we talk about 
kernels on graphs, a few graph-theoretic concepts are needed. 
The {\em adjacency matrix}, 
$\myvec{A}_{\cG}$, for graph $\cG$ is defined as
\[
 \myvec{A}_{\cG}(i,j) = \begin{cases}
 1, & \mbox{if there is an edge between $\myvec{v}_i$ and $\myvec{v}_j$}; \\
 0, & \mbox{if not}.
 \end{cases}
\] 
Let 
\[
d_i = \sum_{j \neq i} \myvec{A}_{\cG}(i, j)
\]
be the {\em degree} of node $\myvec{v}_i$, or simply the number of nodes 
it is adjacent to.
The {\em graph Laplacian matrix} $\myvec{L}_{\cG}$ for $\cG$ is defined by
\[
 \myvec{L}_{\cG} = \myvec{D}_{\cG} - \myvec{A}_{\cG},
 \quad\mbox{where}\quad
 \myvec{D}_{\cG} = \mbox{diag}\{ d_1, d_2, ..., d_n \}.
\]
In other words, 
\[
 \myvec{L}_{\cG}(i,j) = \begin{cases}
   d_i,                 & \mbox{if } i = j; \\
 -\myvec{A}_{\cG}(i,j), & \mbox{if } i \neq j.
 \end{cases} 
\]


With these basic notions, we are now ready to talk about kernels on 
graphs.
For graph data, one often uses a so-called diffusion kernel \citep{diff-ker},
defined as
\beqn
\label{eq:diffKer}
 \myvec{K}_{\cG} = \mbox{exp}(-\beta \myvec{L}_{\cG}) =
 \sum_{m=0}^{\infty} \frac{(-\beta)^m}{m!} \myvec{L}_{\cG}^m,
\eeqn
where $\beta > 0$ is a tuning parameter. 
The meaning and intuition behind the diffusion kernel 
require a relatively long explanation, which we won't go into. Roughly 
speaking, to measure the similarity between 
$\myvec{v}_i$ and $\myvec{v}_j$, the diffusion kernel takes into 
account the number of paths of length $m$ between 
$\myvec{v}_i$ and $\myvec{v}_j$, for all $m$, and gives shorter paths more 
weight \citep[e.g.,][Section 8.4.2]{eric-net}. 

More generally, one is not required to use the adjacency matrix or the 
graph Laplacian to construct the diffusion kernel. Instead, a diffusion 
kernel (\ref{eq:diffKer}) can be constructed as long as $\myvec{L}_{\cG}$ 
is a valid similarity matrix \citep[see, e.g.,][]{kernel-bk}; we will show 
an example below (Section~\ref{sec:lawyer}).

To compute the diffusion kernel, let $\myvec{L}_{\cG} = 
\myvec{U\Sigma}\myvec{U}^T$ be the spectral decomposition 
of the 
Laplacian 
matrix,
where $\myvec{\Sigma} = \mbox{diag}(s_m)$. Using the fact that 
$\myvec{L}_{\cG}^m$ has the same 
eigenvectors 
for all $m$, the diffusion kernel can be computed by
\[
 \myvec{K}_{\cG} = 
 \myvec{U} \mbox{diag}\left(e^{-\beta s_m}\right) \myvec{U}^T. 
\]
We will use $\myvec{K}_{\cG}(i,j)$ and 
$\myvec{K}_{\cG}(\myvec{v}_i,\myvec{v}_j)$ interchangeably.

\subsection{A simple kernel machine}
\label{sec:simple-km}

\def\cF{\mathcal{F}}
\def\kG{\myvec{K}_{\cG}}
\newcommand{\ind}[1]{\substack{\myvec{v}_i \in \cD\\y_i={#1}}}

Since $\kG(\myvec{v}_i,\myvec{v}_j)$ measures the similarity between
nodes $\myvec{v}_i$ and $\myvec{v}_j$,
to classify $\myvec{v}_0 \in \cDn$
we can clearly use the function,
\beqn
\label{eq:kdclassify}
 f_{\cG}(\myvec{v}_0) \equiv
 \frac{1}{n_1} \sum_{\ind{1}}
  \kG \left(\myvec{v}_0, \myvec{v}_i \right)
-
 \frac{1}{n_2} \sum_{\ind{2}}
  \kG \left(\myvec{v}_0, \myvec{v}_i \right),
\eeqn
where
$n_1$ and $n_2$ are total number of nodes in $\cD$ with class label 
$1$ and $2$, respectively.
For $\myvec{v}_0$, this is simply the difference between its
average similarity to class 1 and its average similarity to class 2. 
For example, we can classify $\myvec{v}_0$ to the class which it is 
more similar to, i.e., 
\beqn
\label{eq:kdrule}
 \hat{y}_0 = \begin{cases}
 1, & \mbox{if } f_{\cG}(\myvec{v}_0) > c, \\
 2, & \mbox{if otherwise},
 \end{cases}
\eeqn
for some thresholding constant $c$.  
Below, we will sometimes drop the subscript ``$\myvec{v}_i \in 
\cD$'' from the summation in (\ref{eq:kdclassify}).

It is easy to see that the function $f_{\cG}(\myvec{v}_0)$ in 
(\ref{eq:kdclassify}) is of the general form (\ref{eq:km}), with $\alpha_0 
= 0$, $\alpha_i = 1/n_1$ or $-1/n_2$ depending on whether $y_i = 1$ or 2, 
and $\myvec{K}=\kG$. In other words, (\ref{eq:kdclassify}) is a simple 
kernel machine. We shall mostly work with this simple kernel machine 
because it can easily be constructed without invoking expensive 
optimization procedures in order to determine the coefficients, $\alpha_0, 
\alpha_1, \alpha_2, ..., \alpha_n$. A more sophisticated kernel machine 
such as the SVM, on the other hand, would require quadratic programming to 
find these coefficients. However, though we use (\ref{eq:kdclassify}) for 
convenience, we emphasize the framework we develop below does not preclude 
us from using other kernel machines such as SVMs.

\section{Deep kernel machines on $\cG$}
\label{sec:main}


A key idea behind kernel machines is that
kernels can be regarded as calculating inner products in an
implicit feature space, call it $\cF$. That is,
\[
 \kG(\myvec{v}_i,\myvec{v}_j) = \iprodphiv{i}{j},
 \quad\mbox{where}\quad
 \phi: V(\cG) \mapsto \cF.
\]
This means the kernel $\kG$ necessarily induces a distance function in
$\cF$,
\beqn
 d_{\cF}(\myvec{v}_i, \myvec{v}_j) 
&=& \| \phi(\myvec{v}_i) - \phi(\myvec{v}_j) \| \notag \\
&=& \sqrt{ \iprodphiv{i}{i} - 2\iprodphiv{i}{j} + \iprodphiv{j}{j} } \notag\\
&=& \sqrt{ \kG(\myvec{v}_i,\myvec{v}_i) 
        - 2\kG(\myvec{v}_i,\myvec{v}_j) 
        +  \kG(\myvec{v}_j,\myvec{v}_j) } 
\label{eq:dF}
\eeqn

\subsection{Linear classification in $\cF$}

Using the distance $d_{\cF}$ --- more specificly the squared distance 
$d^2_{\cF}$, the decision rule (\ref{eq:kdrule}) is 
equivalent to nearest-centroid classification in the feature space $\cF$.
To see this, notice that
\[
 \frac{1}{n_1} \sum_{y_i=1} \phi(\myvec{v}_i) 
 \quad\mbox{and}\quad
 \frac{1}{n_2} \sum_{y_i=2} \phi(\myvec{v}_i)
\]
are class centroids in $\cF$. Nearest-centroid classification 
simply declares 
$\hat{y}_0 = 1$ if $\myvec{v}_0$ is closer to the centroid of class 1, 
i.e., if
\[
 \left\| \phi(\myvec{v}_0) -  
  \frac{1}{n_1} \sum_{y_i=1} \phi(\myvec{v}_i) 
  \right\|^2
<
 \left\| \phi(\myvec{v}_0) - 
  \frac{1}{n_2} \sum_{y_i=2} \phi(\myvec{v}_i) 
  \right\|^2.
\]
This is equivalent to
\begin{multline*}
\left[
\iprodphiv{0}{0} - \frac{2}{n_1} \sum_{y_i=1} \iprodphiv{i}{0} +
\frac{1}{n_1^2} \sum_{y_i, y_j=1} \iprodphiv{i}{j} \right] \\ -
\left[
\iprodphiv{0}{0} - \frac{2}{n_2} \sum_{y_i=2} \iprodphiv{i}{0} +
\frac{1}{n_2^2} \sum_{y_i, y_j=2} \iprodphiv{i}{j} \right]
< 0.
\end{multline*}
Cancelling out $\iprodphiv{0}{0}$ and dividing by $-2$, we obtain
\[
 \frac{1}{n_1} \sum_{y_i=1}
  \kG \left(\myvec{v}_0, \myvec{v}_i \right)
-
 \frac{1}{n_2} \sum_{y_i=2}
  \kG \left(\myvec{v}_0, \myvec{v}_i \right) - c(\cD) > 0,
\]
where
\[
 c(\cD) \equiv
 \frac{1}{2 n_1^2} \sum_{y_i, y_j=1} \kG(\myvec{v}_i, \myvec{v}_j) -
 \frac{1}{2 n_2^2} \sum_{y_i, y_j=2} \kG(\myvec{v}_i, \myvec{v}_j)  
\]
is a constant that depends only on $\cD$ and not on $\myvec{v}_0$. 
Clearly, this is equivalent to (\ref{eq:kdrule}).  

Being equivalent to nearest-centroid classification, the 
kernel machine (\ref{eq:kdclassify}) is a linear classifier in the feature 
space $\cF$. In fact, most kernel machines are linear in the 
feature space, including SVMs.

\subsection{Nonlinear classification in $\cF$}

\def\kF{\myvec{K}_{\cF}}

However, it is quite possible that a linear classifier in $\cF$ is not 
sufficient; we will show some examples below (Section~\ref{sec:ex}). But, 
in 
principle, there is nothing that prevents us from using 
other, more flexible 
classifiers in $\cF$. For example, using the distance $d_{\cF}$, we may 
consider a classifier based on kernel density estimates. 
Let 
\beqn
\label{eq:kde}
 \hat{p}_k(\myvec{v}) = 
 \frac{1}{n_l} \sum_{\substack{\myvec{v}_i \in \cD\\y_i=k}}
 K_{h({\cF})} \left( d_{\cF}(\myvec{v}, \myvec{v}_i) 
     \right)
\eeqn
be a kernel density estimate of the distribution for class $k$. 
Many kernel functions can be used for density estimation, e.g., 
\beqn
\label{eq:rbKer}
K_{h({\cF})}(d_{\cF}) = \frac{1}{\sqrt{2\pi} h({\cF})} \mbox{exp}
\left( -\frac{d_{\cF}^2}{2h^2({\cF})} \right), 
\eeqn
where $h({\cF})$ is a bandwidth parameter, which serves to scale the 
distance $d_{\cF}$. We shall write
\beqn
\label{eq:kF}
\kF(\myvec{v}_i,\myvec{v}_j) 
&\equiv& K_{h({\cF})} \left( d_{\cF}(\myvec{v}_i,\myvec{v}_j) \right).
\eeqn
Using (\ref{eq:rbKer}) for $K_{h({\cF})}$, $\kF$ is nothing but the 
well-known 
radial-basis or Gaussian kernel, except it uses the
distance function $d_{\cF}$ rather than a distance defined on the 
original graph $\cG$. Therefore,
(\ref{eq:kde}) is a density estimate in the space $\cF$ rather than on 
the original graph $\cG$. 
The subscript $\cF$ and the notation $h(\cF)$
 are used to emphasize this fact and to 
differentiate $\kF$ from $\kG$, the kernel on the original graph $\cG$ 
that induced the space $\cF$. 

Based on the kernel density estimates in 
(\ref{eq:kde}), for 
each $\myvec{v}_0 \in \cDn$ we can predict its class label 
$\hat{y}_0$ depending on whether
$\hat{p}_1(\myvec{v}_0) - \hat{p}_2(\myvec{v}_0)$ is positive or negative.
In other words, the decision function is simply
\beqn
\label{eq:kdclassifyF}
 f_{\cF}(\myvec{v}_0) = 
 \frac{1}{n_1} \sum_{\ind{1}}
 \kF \left(\myvec{v}_0, \myvec{v}_i \right)
-
 \frac{1}{n_2} \sum_{\ind{2}}
 \kF \left(\myvec{v}_0, \myvec{v}_i \right).
\eeqn
It is easy to see that (\ref{eq:kdclassifyF}) is another kernel machine of 
the same form as (\ref{eq:kdclassify}); the only difference is that 
(\ref{eq:kdclassifyF}) uses the kernel $\kF$ whereas (\ref{eq:kdclassify}) 
uses the kernel $\kG$. 

\subsection{Deep kernel machines}

Let us summarize what we have said so far. The space $\cF$ is the implicit 
feature space for $\kG$. A kernel machine $f_{\cG}$ (\ref{eq:kdclassify}) 
using the kernel $\kG$ is a linear classifier in $\cF$. If linear 
classifiers are not sufficient in $\cF$, we can relax linearity and choose 
to work with a nonlinear classifier, e.g., by constructing kernel density 
estimates in $\cF$ via the implied distance metric $d_{\cF}(\myvec{v}_i, 
\myvec{v}_j)$ --- equation~(\ref{eq:dF}). This gives rise to a new kernel 
machine $f_{\cF}$ (\ref{eq:kdclassifyF}), using the kernel 
$\kF(\myvec{v}_i,\myvec{v}_j)$ --- equation~(\ref{eq:kF}). If we use 
(\ref{eq:rbKer}) for kernel density estimation, the kernel updating 
formula from $\kG$ to $\kF$ is simply, putting (\ref{eq:dF}), 
(\ref{eq:rbKer}), and (\ref{eq:kF}) together,
\beqn
\label{eq:kFspec}
\kF(\myvec{v}_i,\myvec{v}_j) 
&=& 
\frac{1}{\sqrt{2\pi} h({\cF})} \mbox{exp}
\left( -\frac{\kG(\myvec{v}_i,\myvec{v}_i) 
            -2\kG(\myvec{v}_i,\myvec{v}_j) 
            + \kG(\myvec{v}_j,\myvec{v}_j)}{2h^2({\cF})} 
\right).
\eeqn
The choice of $h({\cF})$ will be discussed below 
(Section~\ref{sec:heuristic}).

However, there is no reason why the process must end here. The kernel 
$\kF$ has its implicit feature space as well; let's  
call it 
$\cF^2$. The kernel machine $f_{\cF}$ (\ref{eq:kdclassifyF}) using the 
kernel $\kF$ is a linear classifier in $\cF^2$. 
We can relax linearity in $\cF^2$, if necessary, and choose to work with 
a nonlinear classifier, again, by constructing kernel density 
estimates in $\cF^2$ via the implied distance metric,
\[
 d_{\cF^2}(\myvec{v}_i, \myvec{v}_j) = 
    \sqrt{ \kF(\myvec{v}_i,\myvec{v}_i) 
        - 2\kF(\myvec{v}_i,\myvec{v}_j) 
        +  \kF(\myvec{v}_j,\myvec{v}_j) }. 
\]
By the same argument, this would give us yet another kernel machine, say 
$f_{\cF^2}$, of exactly  
the same form as $f_{\cG}$ (\ref{eq:kdclassify}) and $f_{\cF}$ 
(\ref{eq:kdclassifyF}), except it would be using the kernel
\beqnn
\myvec{K}_{\cF^2}(\myvec{v}_i,\myvec{v}_j) 
&\equiv& K_{h({\cF^2})} \left( d_{\cF^2}(\myvec{v}_i,\myvec{v}_j) \right) \\
&=& 
\frac{1}{\sqrt{2\pi} h({\cF^2})} \mbox{exp}
\left( -\frac{\kF(\myvec{v}_i,\myvec{v}_i) 
            -2\kF(\myvec{v}_i,\myvec{v}_j) 
            + \kF(\myvec{v}_j,\myvec{v}_j)}{2h^2({\cF^2})} 
\right).
\eeqnn

It is easy to see that this process can be repeated recursively (see 
Table~\ref{tab:dkm}). We refer to kernel machines generated by this 
recursive process as ``deep kernel machines'' (DKMs). The one using the 
original kernel $\kG$ is a referred to as a level-0 DKM; the one using the 
kernel $\kF$, a level-1 DKM; the one using the kernel $\myvec{K}_{\cF^2}$, 
a level-2 DKM; and so on. Notice that the DKM algorithm presented in 
Table~\ref{tab:dkm} is slightly more general than what we have discussed 
above. In our discussion, we have focused on a specific base kernel 
machine (Section~\ref{sec:simple-km}), but one can certainly use other 
base kernel machines, e.g., SVMs (see Section~\ref{sec:bkms} below).

\btab
\centering
\mycap{\label{tab:dkm}%
Pseudo code for deep kernel machines (DKMs).}
\fbox{%
\begin{tabular}{p{0.85\textwidth}}
{\bf function} BaseKernelMachine($\myvec{K}$, $\cD$, $\cDn$) \\
for (every $\myvec{v} \in \cDn$) \{ \\
\[
 f(\myvec{v}) = \alpha_0 + 
 \sum_{\myvec{v}_i \in \cD} \alpha_i \myvec{K}(\myvec{v}, \myvec{v}_i),
\]
~~e.g., $\alpha_0=0$, $\alpha_i = 1/n_1$ if $y_i=1$ 
and $\alpha_i = -1/n_2$ if $y_i=2$. \\
\} \\
return $f$; \\
{\bf end function} \\

\vspace{3mm}

{\bf function} GetKernel($\kG$, $level$) \\
if ($level == 0$) \{ \\ 
~~~return $\kG$; \\
\} \\
else \{ \\
~~~compute $d_{\cF}$ according to equation~(\ref{eq:dF}):
\[ 
  d_{\cF}(\myvec{v}_i,\myvec{v}_j) = 
  \sqrt{\kG(\myvec{v}_i,\myvec{v}_i)
      -2\kG(\myvec{v}_i,\myvec{v}_j)
       +\kG(\myvec{v}_j,\myvec{v}_j)};
\]
~~~choose $h({\cF})$ according to equation~(\ref{eq:bandchoice}):
\[
  h({\cF}) = \frac{1}{n^2} 
  \underset{\myvec{v}_i,\myvec{v}_j \in \cD}{\sum\sum} 
  d_{\cF}(\myvec{v}_i,\myvec{v}_j);
\]
~~~compute $\kF$ according to equations (\ref{eq:kF}) and (\ref{eq:rbKer}):
\[
 \kF(\myvec{v}_i,\myvec{v}_j) = 
 K_{h({\cF})}\left(d_{\cF}(\myvec{v}_i,\myvec{v}_j)\right);
\]
~~~return GetKernel($\kF$, $level-1$); \\
\} \\
{\bf end function} \\

\vspace{3mm}

{\bf function} DeepKernelMachine($\kG$, $\cD$, $\cDn$, $level$) \\
return BaseKernelMachine(GetKernel($\kG$, $level$), $\cD$, $\cDn$); \\
{\bf end function}
\end{tabular}}
\etab

\subsection{A heuristic for choosing $h({\cF})$}
\label{sec:heuristic}

To carry out density estimation in $\cF$, a bandwidth parameter $h({\cF})$ 
must be specified; see (\ref{eq:kde}) and (\ref{eq:rbKer}). While users 
are certainly free to optimize this parameter in practice, this can be 
tedious for DKMs because, as we go from $\cG$ to $\cF, \cF^2, \cF^3, ...$, 
there is a bandwidth parameter for each space, $h(\cF), h(\cF^2), 
h(\cF^3), ...$, so a heuristic is desired.
A reasonable heuristic is:
\beqn
\label{eq:bandchoice}
 h({\cF}) = \frac{1}{n^2}
     \underset{\myvec{v}_i,\myvec{v}_j \in V(\cG)}{\sum\sum} 
     d_{\cF}(\myvec{v}_i, \myvec{v}_j). 
\eeqn
That is, $h({\cF})$ can be chosen to be the average pair-wise distance in 
the space of $\cF$. We use this heuristic in all of our experiments below.

\section{Experiments}
\label{sec:ex}

In this section, we describe a few experiments and show that DKMs are 
useful. 

\subsection{Enron email data}
\label{sec:enron}

In 2001, a large USA-based gas and electricity company named Enron was 
found guilty of serious accounting frauds, a case that caught worldwide 
attention. As part of the investigation, the US Federal Energy Regulatory 
Commission confiscated its corporate email database and made it publicly 
available. Preprocessed versions of these data can be obtained from 
\url{http://cis.jhu.edu/~parky/Enron/enron.html}. In particular, there is 
a $184 \times 184$ adjacency matrix, $\myvec{A}_{\cG}$, that indicates 
whether there was email communication between any two of 184 unique email 
accounts. Our initial $\kG$ is simply a diffusion kernel 
(\ref{eq:diffKer}) based on this adjacency matrix, but we removed two 
accounts that never sent an email to another account. The status of these 
184 email account owners are also available, which we use to create two 
different classification tasks (see Table~\ref{tab:enron-data}).

\btab
\centering
\mycap{\label{tab:enron-data}%
Part of Enron email data used for tasks 1 and 2. We removed two 
accounts that never sent an email to another account.}
\fbox{%
\begin{tabular}{l|r|cc}
 & & \multicolumn{2}{c}{Class Label} \\
Node Status & $N$ & Unbalanced & Balanced \\
\hline
CEO       & 5  & 1 & 1 \\
President & 5  & 1 & 1 \\
Managing Director & 6 & 1 & 1 \\
Director        & 14  & 2 & 1 \\    
Vice President  & 30  & 2 & 1 \\
Manager   & 16  & 2 & 1 \\
Lawyer    & 1  & 2 & 1 \\
Employee  & 40  & 2 & 2  \\ 
Trader    & 11  & 2 & 2  \\
Other     & 54  & 2 & 2  \\
\hline
Total     & 182 & 16 vs 166 & 77 vs 105 \\
          &     & (Task 1)  & (Task 2) \\

\end{tabular}}
\etab

\subsection{Lazega lawyer data}
\label{sec:lawyer}

\citet{lazega} studied collaborative working relationships and social 
interactions among members of a New England law firm. There were 36 
partners in the firm. The 36 partners were interviewed and asked to 
express their opinions on various issues regarding how the law firm should 
be managed. One of the issues had to do with workflow inside the firm 
\citep[][Chapter 8]{lazega}. Some favored the status quo ($y_i=1$) while 
others favored less flexible workflow ($y_i = 2$). As our third 
classification task (see Table~\ref{tab:lazega-data}), we try to predict 
the partners' position on this 
particular issue based on their social interactions and working 
relationships, e.g., whether any two partners worked together or 
considered themselves to be friends.
Our initial $\kG$ is a diffusion kernel (\ref{eq:diffKer}) based on 
a similarity (rather than adjacency) matrix
$\myvec{A}_{\cG}$, defined as follows:
\[
 \myvec{A}_{\cG}(i,j) = 0.5 \times \myvec{I}_{friends}(i,j) + 
                        0.5 \times \myvec{I}_{collaborated}(i,j),
\]  
where $\myvec{I}_{friends}(i,j) = 1$ if $\myvec{v}_i$ and $\myvec{v}_j$ 
were friends and $0$ if not; and likewise for $\myvec{I}_{collaborated}(i,j)$. 

\btab
\centering
\mycap{\label{tab:lazega-data}%
Part of Lazega lawyer data used for task 3.}
\fbox{%
\begin{tabular}{l|r|c}
Node Position & $N$ & Class Label\\
\hline
Favors status quo               & 20 & 1 \\
Favors less flexible workflow   & 16 & 2 \\
\hline
Total     & 36 & 20 vs 16 \\
          &    & (Task 3) \\
\end{tabular}}
\etab

\subsection{Performance measure}
\label{sec:perf}

Classification task 1 (Table~\ref{tab:enron-data}) is a highly unbalanced 
problem. For this task, we use the average precision 
\citep[e.g.,][Appendix A]{lago}, or simply AP, to evaluate performance. 
The AP is a widely used criterion in the information retrieval community, 
and is particularly suitable for unbalanced classification problems. 
Classification tasks 2 and 3 (Tables \ref{tab:enron-data} and 
\ref{tab:lazega-data}) are relatively balanced problems. For these two 
tasks, we use the area under the receiver-operating characteristic (ROC) 
curve \citep{rocbk}, or simply AUC (for ``area under the curve''), to 
evaluate performance. The main reason for using the AUC and the AP (rather 
than, e.g., total misclassification error) is because they are {\em not} 
affected by the thresholding constant $c$ in equation~(\ref{eq:kdrule}).
Table~\ref{tab:variety} summarizes the main features of the three tasks. 

\btab
\centering
\mycap{\label{tab:variety}%
Summary of classification tasks.}
\fbox{%
\begin{tabular}{l|r|c|c|c}
    & \multicolumn{1}{c|}{\%}        & Data & & Performance \\
    & \multicolumn{1}{c|}{($y_i=1$)} & Set & $\myvec{A}_{\cG}$  & Measure \\
\hline
Task 1 & 8.8~~  & Enron & adjacency matrix & AP \\
Task 2 & 42.3~~ & Enron & adjacency matrix & AUC \\
Task 3 & 55.6~~ & Lawyer & custom similarity matrix & AUC \\
\end{tabular}}
\etab

\subsection{Base kernel machines}
\label{sec:bkms}

We use two types of base kernel machines to run DKMs: the simple kernel 
machine (\ref{eq:kdclassify}) and the SVM. Both are linear in the feature 
space, but SVM directly goes after the optimal hyperplane. Notice that, to 
fit an SVM, one must specify the amount of penalty on the sum of slack 
variables, often called the ``cost'' parameter in most SVM packages. Every 
time an SVM is fitted, we simply choose the best ``cost'' parameter among 
a wide range of values:
\[
10^{-5}, 10^{-4}, 10^{-3}, 10^{-2}, 0.05, 0.1, 0.5, 1, 2, 5, 10, 50, 100. 
\]
This gives the SVM an unfair advantage, but, since we are using SVMs as a 
benchmark, it is well understood that giving the benchmark an unfair 
advantage will only lead us to more conservative conclusions. In reality, 
this extra ``cost'' parameter in the SVM must be chosen by cross 
validation on $\cD$. 

\subsection{Results and conclusions}

Figure~\ref{fig:rslt} shows the average results over 25 random splits of 
$V(\cG)$ into $\cD$ and $\cDn$. The random splits are stratified by class 
label so that the fraction of nodes belonging to each class is roughly the 
same on both $\cD$ and $\cDn$. The main conclusions we can draw from 
Figure~\ref{fig:rslt} are as follows:

\bitem

\item[(C1)] Though it goes directly after an ``optimal'' linear 
classifier, the SVM is not necessarily a better base kernel machine than a 
simple kernel machine such as (\ref{eq:kdclassify}). Both are linear in 
the feature space. It is more important to be in the ``right'' feature 
space than to use an ``optimal'' linear classifier. Using the ``optimal'' 
linear classifier in the ``wrong'' feature space is not going to give you 
good results. DKMs address this issue directly by providing a recursive 
algorithm to look for the ``right'' feature space.

\item[(C2)] When the initial kernel $\kG$ is badly specified, e.g., if the 
tuning parameter $\beta$ is not well chosen for the underlying prediction 
task, DKMs can often boost up the performance significantly. This shows 
that DKMs have an attractive ``automatic kernel correction'' capability. 
When linear classification in the initial feature space $\kF$ is not 
enough to produce good results, it often pays to relax linearity and to go 
up to higher-level feature spaces. DKMs provide an automatic way to do so.

\eitem

\bfig
\centering
\includegraphics[width=0.4\textwidth, angle=270]{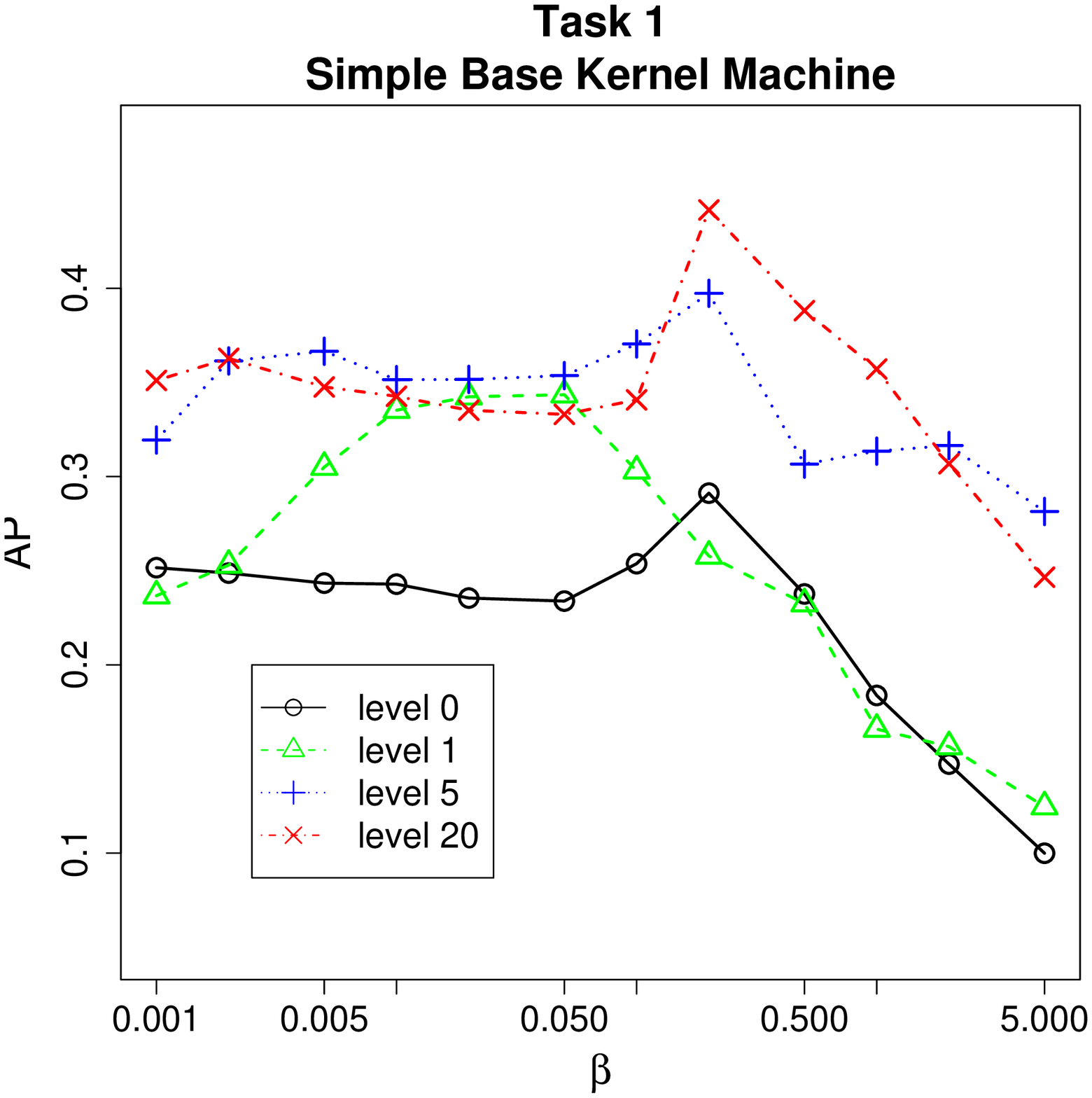}
\includegraphics[width=0.4\textwidth, angle=270]{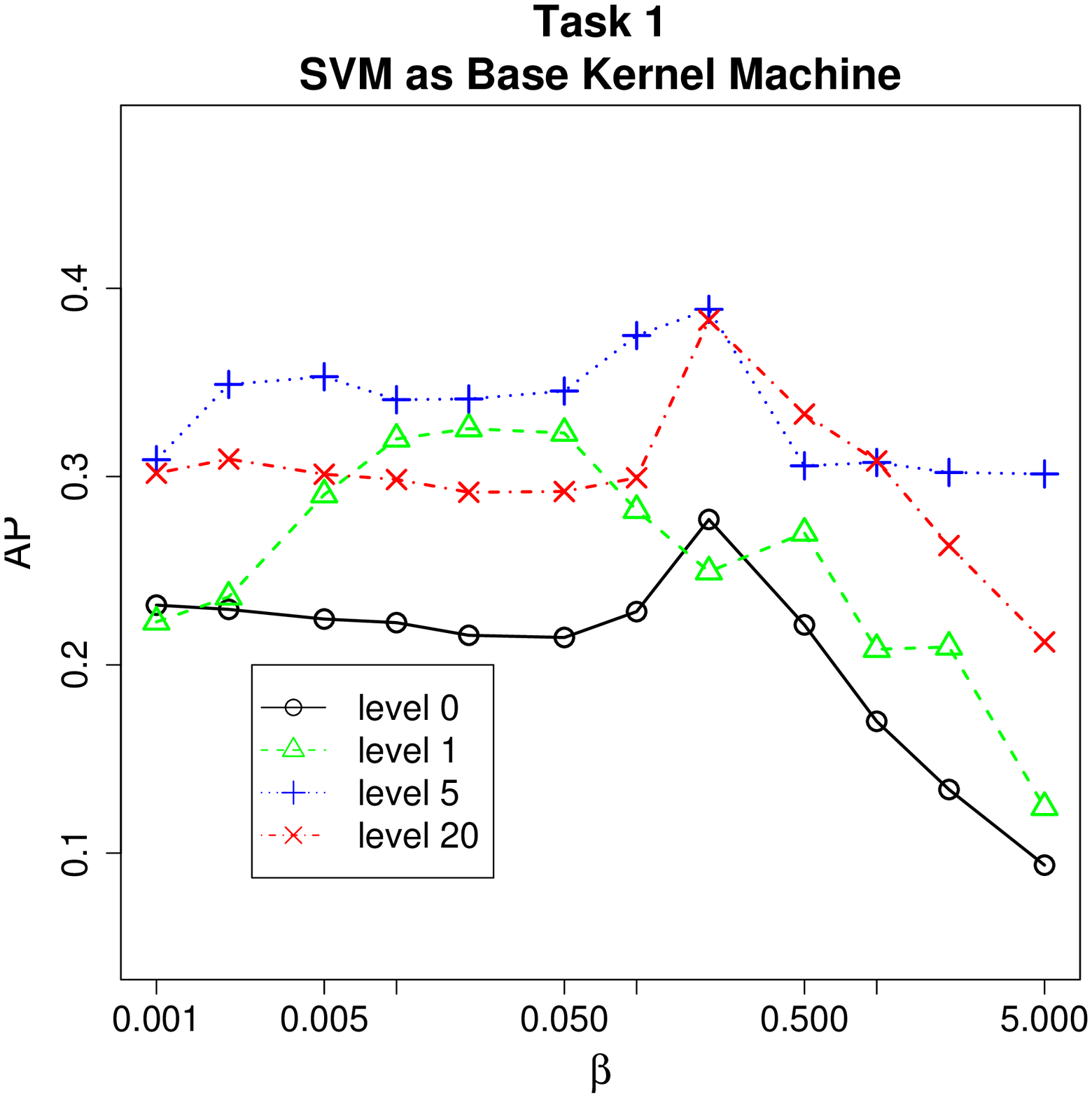} \\
\includegraphics[width=0.4\textwidth, angle=270]{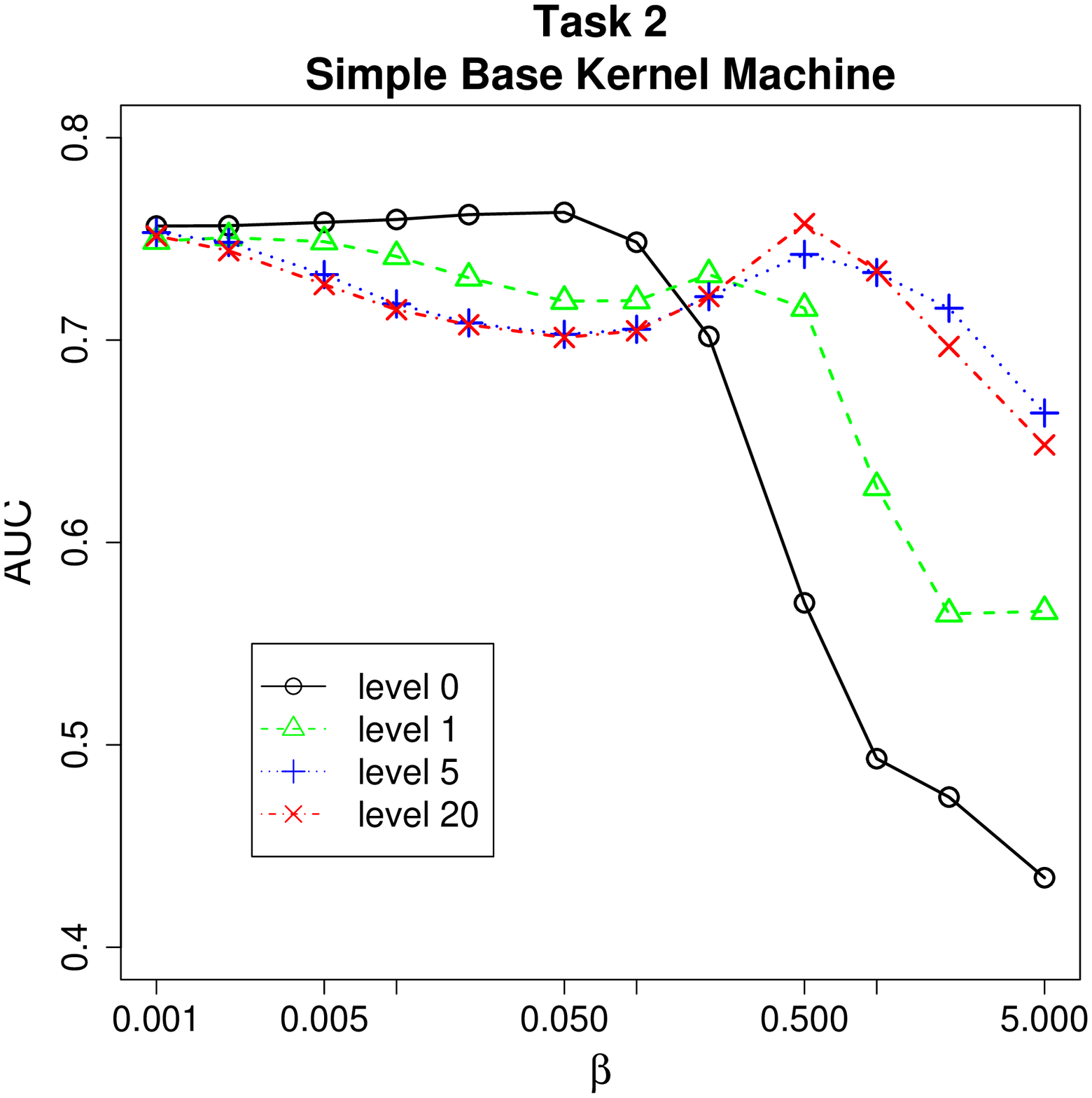}
\includegraphics[width=0.4\textwidth, angle=270]{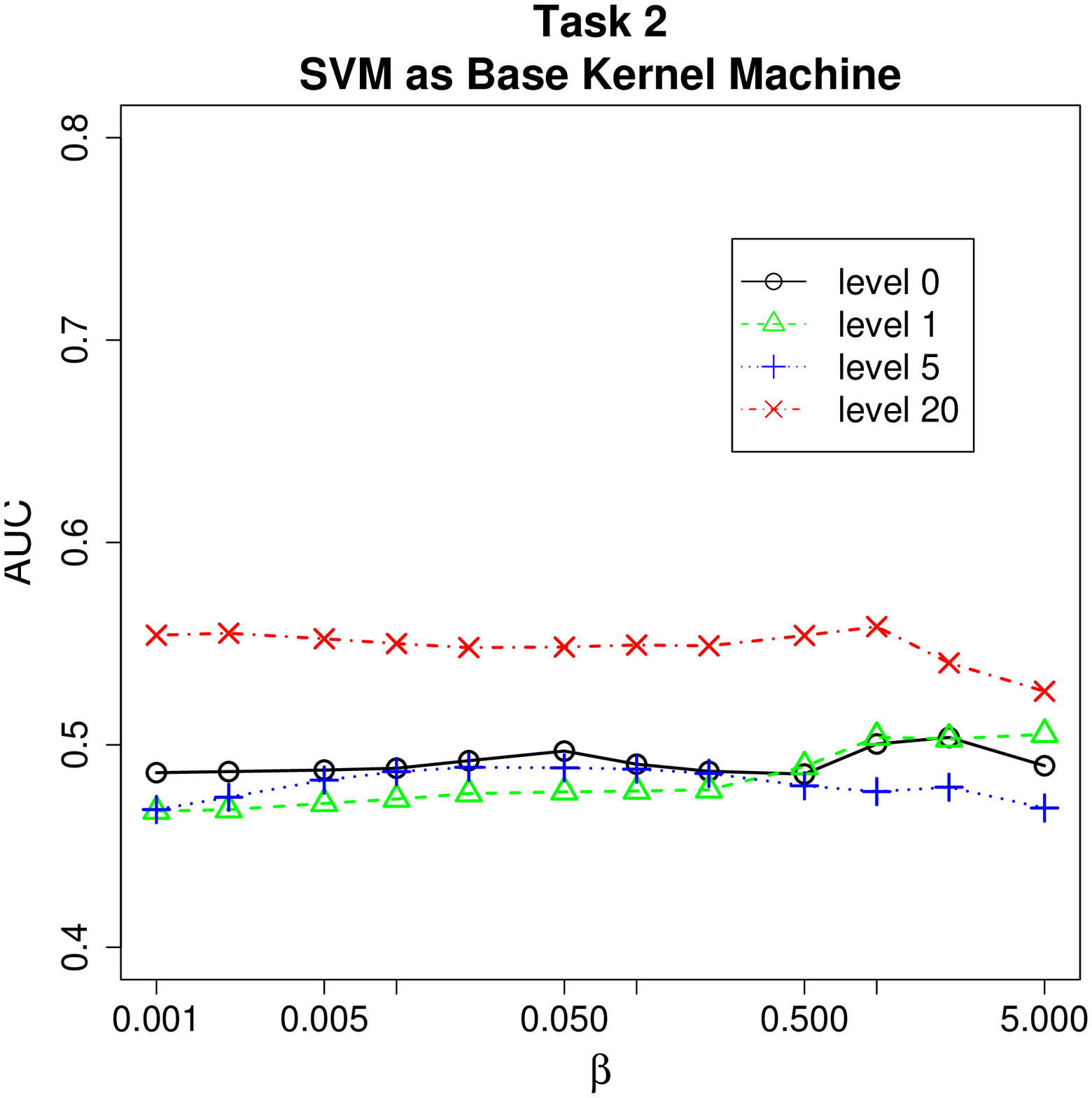} \\
\includegraphics[width=0.4\textwidth, angle=270]{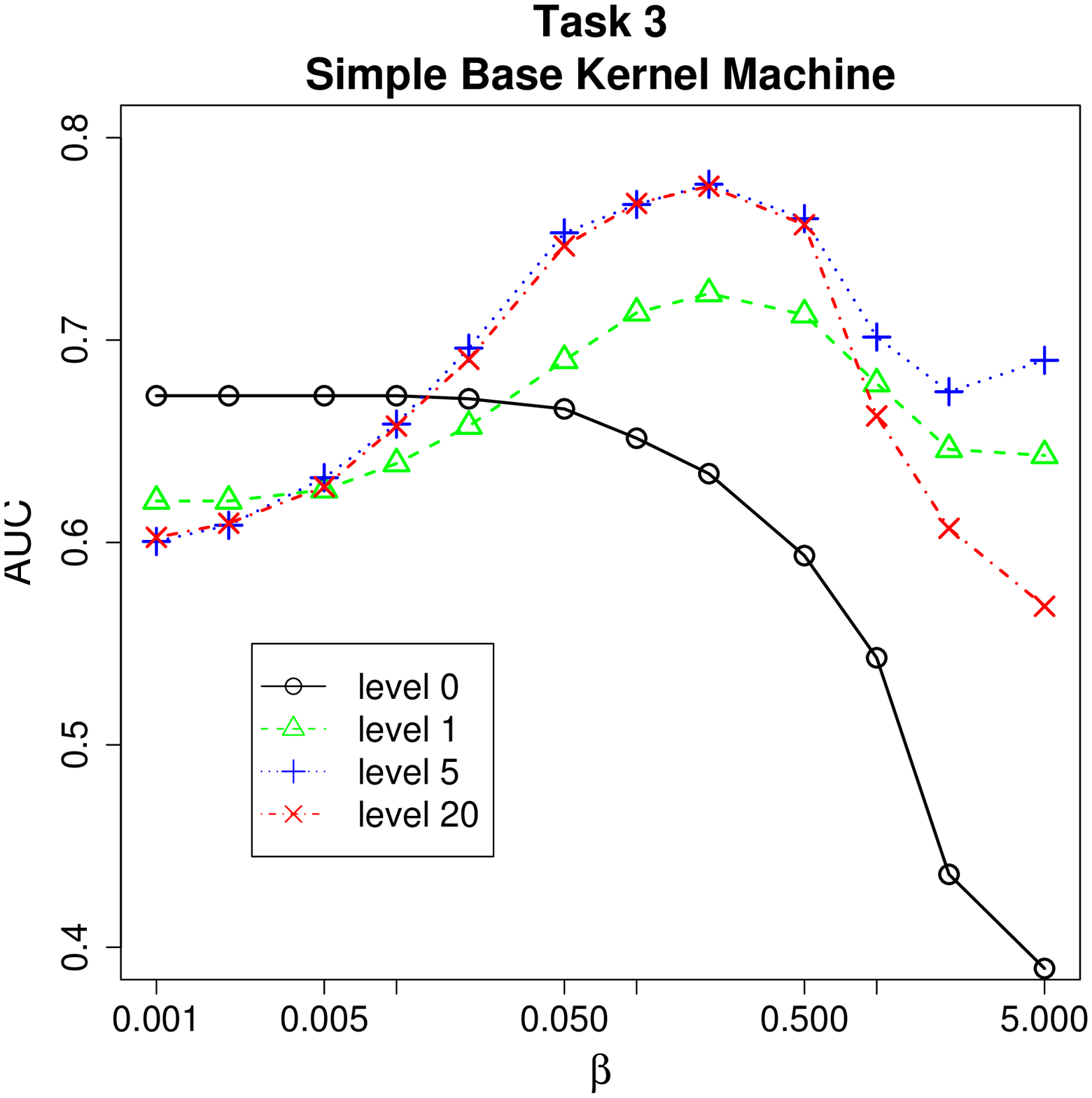}
\includegraphics[width=0.4\textwidth, angle=270]{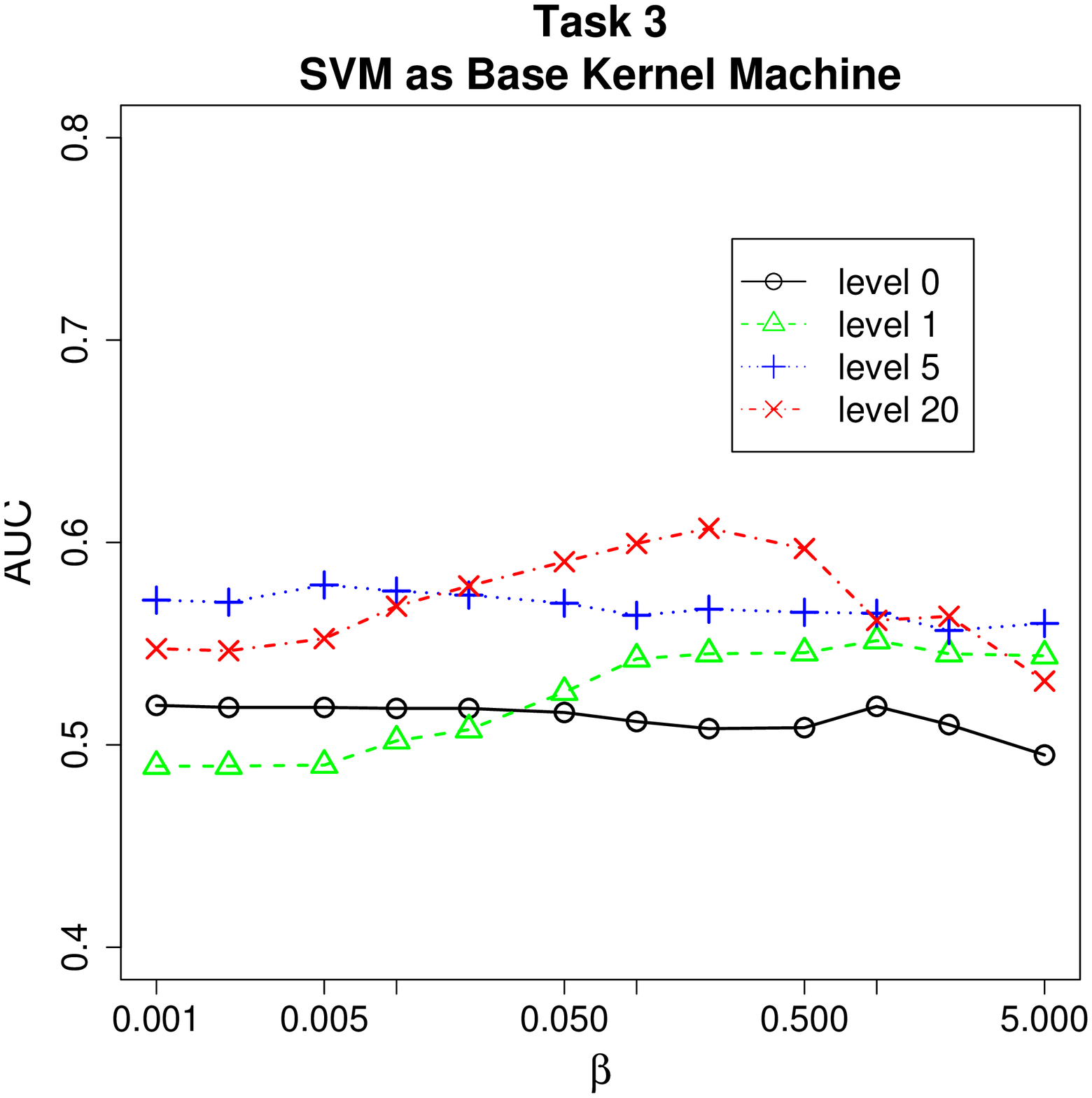}
\mycap{Average performance over 25 random splits of $V(\cG)$ 
into $\cD$ and $\cDn$. 
Horizontal axis (logarithmic scale): tuning parameter $\beta$ for the 
initial diffusion kernel
$\kG$; see equation~(\ref{eq:diffKer}). Vertical axis: performance 
measure, e.g., AP or AUC; see Section~\ref{sec:perf}.
}  
\label{fig:rslt}
\efig

\section{Discussion}
\label{sec:disc}

To paraphrase our conclusions (C1) and (C2) above, we have essentially 
argued that we should put more emphasis on finding the ``right'' feature 
space rather than finding the ``optimal'' linear classifier (perhaps in 
the ``wrong'' feature space). One can use DKMs to do this. While the 
automatic and recursive kernel correction formula (\ref{eq:kFspec}) is 
attractive, there clearly remains one important question that we haven't 
quite addressed: how deep should we go?

Before we address this question, we first briefly mention interesting 
connections between our work and some recent literature on deep 
architectures in machine learning. The neural network was a leading 
algorithm for machine learning during the 1980's, but it did not enjoy as 
wide a success as was initially anticipated. The main reason is because 
the back propagation algorithm is not practical for training neural 
networks that are more than a few layers deep. Recently, many arguments 
\citep[e.g.,][]{deep-nnet-science, deep-beliefnet, 
deep-architecture-bengio} have been made that deep neural networks (i.e., 
neural networks with many layers) are necessary, and practically realistic 
algorithms have also emerged \citep[e.g.,][]{deep-beliefnet-algo, 
deep-nnet-bengio}. Our work provides further support for the idea of deep 
architectures.

By definition, the architecture of a deep neural network is necessarily 
complex. One has to make many decisions. How many layers? How many hidden 
components for each layer? In the landmark article 
\citep{deep-nnet-science} on deep neural networks that appeared in the 
prestigious journal, {\em Science}, the authors showcased deep neural nets 
for a number of different tasks. A very striking feature of that article 
is that the authors used {\em vastly} different deep architectures for the 
different tasks, but there was little explanation on how those 
architectural decisions were made. We asked the first author of the {\em 
Science} article in person, after he delivered a seminar on the very 
subject. His answer was: one simply tries different architectures and 
picks the one that gives the best results. While this is not entirely 
satisfactory, we think such a limitation alone is no reason for anyone to 
deny that deep neural networks are a major advance in modern machine 
learning research. One cannot solve all problems at once. New ideas always 
lead to new problems, and that's the very nature of scientific research.

At this moment, we don't have an entirely satisfactory answer to the 
question of how deep a DKM one should use, except that we have empirically 
observed diminishing returns as we go to higher and higher levels, but 
this limitation alone in our work is no reason for us to reject the fact 
that DKMs can be quite useful. 

Finally, it is not hard to see that the development of these DKMs 
(Section~\ref{sec:main}) does not rely on $\cG$ being a graph. For 
example, if we abuse our notation and allow $\cG$ to denote the usual 
$q$-dimensional Euclidean space, then we simply have a usual 
classification problem --- $\cD$ simply becomes the set of training data 
and $\cDn$, the set of unlabelled observations to be classified. Of 
course, in that case $\kG$ will no longer be the diffusion kernel, but, 
regardless of what it is, a level-0 kernel machine using $\kG$ will still 
be linear in its implicit feature space $\cF$. Using the distance function 
$d_{\cF}$, we can still do kernel density estimation in the space of 
$\cF$, and obtain a level-1 kernel machine using a new kernel $\kF$. In 
other words, the idea of DKMs is general and not restricted to node 
classification on graphs. Whether they are actually useful for data 
structures other than graphs remains to be seen. We leave this to further 
investigation.

\section{Summary}
\label{sec:summary}

We have described the idea of using deep kernel machines for node 
classification on graphs. We have conducted a few experiments to show that 
linear classification in the implicit feature space of kernels commonly 
used for graph data (e.g., the diffusion kernel) is often not enough. When 
this is the case, one can apply the ``kernel trick'' again in the implicit 
feature space itself. Repeating this process leads to deep kernel machines 
(DKMs). Our experiments have shown that DKMs' recursive, automatic kernel 
correction capability is especially useful when the initial kernel $\kG$ 
is not well specified. While the work we reported here is just a beginning 
and there remains much to be done, our results lend support to the idea of 
using deep architectures for machine learning that has recently emerged in 
the literature.